\documentclass[10pt,twocolumn,letterpaper]{article}

\usepackage[pagenumbers]{cvpr} %

\usepackage[dvipsnames,table]{xcolor}

\usepackage[linesnumbered,ruled,vlined]{algorithm2e}
\usepackage{algpseudocode}
\usepackage{comment}
\SetKwInput{KwInput}{Input} 

\SetCommentSty{mycommfont}
\usepackage{amssymb}
\usepackage{array}
\usepackage{mwe}
\usepackage{multirow} 
\usepackage{adjustbox}
\usepackage{epstopdf}
\usepackage{stackengine}
\usepackage{marvosym}

\usepackage{pifont}
\usepackage{makecell}
\usepackage{floatrow}
\newfloatcommand{capbtabbox}{table}[][\FBwidth]

\newcommand{\condenseparagraph}[1]{\noindent\textbf{#1}\quad}

\definecolor{car}{rgb}{0.39215686, 0.58823529, 0.96078431}
\definecolor{bicycle}{rgb}{0.39215686, 0.90196078, 0.96078431}
\definecolor{motorcycle}{rgb}{0.11764706, 0.23529412, 0.58823529}
\definecolor{truck}{rgb}{0.31372549, 0.11764706, 0.70588235}
\definecolor{other-vehicle}{rgb}{0.39215686, 0.31372549, 0.98039216}
\definecolor{person}{rgb}{1.        , 0.11764706, 0.11764706}
\definecolor{bicyclist}{rgb}{1.        , 0.15686275, 0.78431373}
\definecolor{motorcyclist}{rgb}{0.58823529, 0.11764706, 0.35294118}
\definecolor{road}{rgb}{1.        , 0.        , 1.        }
\definecolor{parking}{rgb}{1.        , 0.58823529, 1.        }
\definecolor{sidewalk}{rgb}{0.29411765, 0.        , 0.29411765}
\definecolor{other-ground}{rgb}{0.68627451, 0.        , 0.29411765}
\definecolor{building}{rgb}{1.        , 0.78431373, 0.        }
\definecolor{fence}{rgb}{1.        , 0.47058824, 0.19607843}
\definecolor{vegetation}{rgb}{0.        , 0.68627451, 0.        }
\definecolor{trunk}{rgb}{0.52941176, 0.23529412, 0.        }
\definecolor{terrain}{rgb}{0.58823529, 0.94117647, 0.31372549}
\definecolor{pole}{rgb}{1.        , 0.94117647, 0.58823529}
\definecolor{traffic-sign}{rgb}{1.        , 0.        , 0.    }   
\definecolor{other-struct}{rgb}{1., 0.58823529411, 0}
\definecolor{other-object}{rgb}{0.19607843137, 1., 1.}

\makeatletter
\newcommand{\car@semkitfreq}{3.92}
\newcommand{\bicycle@semkitfreq}{0.03}
\newcommand{\motorcycle@semkitfreq}{0.03}
\newcommand{\truck@semkitfreq}{0.16}
\newcommand{\othervehicle@semkitfreq}{0.20}
\newcommand{\person@semkitfreq}{0.07}
\newcommand{\bicyclist@semkitfreq}{0.07}
\newcommand{\motorcyclist@semkitfreq}{0.05}
\newcommand{\road@semkitfreq}{15.30}  %
\newcommand{\parking@semkitfreq}{1.12}
\newcommand{\sidewalk@semkitfreq}{11.13}  %
\newcommand{\otherground@semkitfreq}{0.56}
\newcommand{\building@semkitfreq}{14.10}  %
\newcommand{\fence@semkitfreq}{3.90}
\newcommand{\vegetation@semkitfreq}{39.30}  %
\newcommand{\trunk@semkitfreq}{0.51}
\newcommand{\terrain@semkitfreq}{9.17} %
\newcommand{\pole@semkitfreq}{0.29}
\newcommand{\trafficsign@semkitfreq}{0.08}
\newcommand{\semkitfreq}[1]{{\csname #1@semkitfreq\endcsname}}

\newcommand{\car@kittithreesixtyfreq}{2.85}
\newcommand{\bicycle@kittithreesixtyfreq}{0.02}
\newcommand{\motorcycle@kittithreesixtyfreq}{0.01}
\newcommand{\truck@kittithreesixtyfreq}{0.16}
\newcommand{\othervehicle@kittithreesixtyfreq}{0.58}
\newcommand{\person@kittithreesixtyfreq}{0.02}
\newcommand{\road@kittithreesixtyfreq}{14.98}  %
\newcommand{\parking@kittithreesixtyfreq}{2.31}
\newcommand{\sidewalk@kittithreesixtyfreq}{6.43}  %
\newcommand{\otherground@kittithreesixtyfreq}{2.05}
\newcommand{\building@kittithreesixtyfreq}{15.67}  %
\newcommand{\fence@kittithreesixtyfreq}{0.96}
\newcommand{\vegetation@kittithreesixtyfreq}{41.99}  %
\newcommand{\terrain@kittithreesixtyfreq}{7.10} %
\newcommand{\pole@kittithreesixtyfreq}{0.22}
\newcommand{\trafficsign@kittithreesixtyfreq}{0.06}
\newcommand{\otherstruct@kittithreesixtyfreq}{4.33}
\newcommand{\otherobject@kittithreesixtyfreq}{0.28}
\newcommand{\kittithreesixtyfreq}[1]{{\csname #1@kittithreesixtyfreq\endcsname}}

\definecolor{cvprblue}{rgb}{0.21,0.49,0.74}
\usepackage[pagebackref,breaklinks,colorlinks,citecolor=cvprblue]{hyperref}

\usepackage[capitalize]{cleveref}

\title{Real-Time 3D Occupancy Prediction via Geometric-Semantic Disentanglement}

\author{
	Yulin He$^1$ \quad 
        Wei Chen$^{1}\textsuperscript{\Letter}$  \quad 
        Tianci Xun$^{1}$  \quad 
        Yusong Tan$^{1}$  \quad \\
	$^1$National University of Defense Technology\vspace{0.2cm}\\
}

\begin{document}
\maketitle

\begin{abstract}
Occupancy prediction plays a pivotal role in autonomous driving (AD) due to the fine-grained geometric perception and general object recognition capabilities.
However, existing methods often incur high computational costs, which contradicts the real-time demands of AD.
To this end, we first evaluate the speed and memory usage of most public available methods, aiming to redirect the focus from solely prioritizing accuracy to also considering efficiency.
We then identify a core challenge in achieving both fast and accurate performance: \textbf{the strong coupling between geometry and semantic}.
To address this issue, 1) we propose a Geometric-Semantic Dual-Branch Network (GSDBN) with a hybrid BEV-Voxel representation.
In the BEV branch, a BEV-level temporal fusion module and a U-Net encoder is introduced to extract dense semantic features.
In the voxel branch, a large-kernel re-parameterized 3D convolution is proposed to refine sparse 3D geometry and reduce computation.
Moreover, we propose a novel BEV-Voxel lifting module that projects BEV features into voxel space for feature fusion of the two branches.
In addition to the network design, 2) we also propose a Geometric-Semantic Decoupled Learning (GSDL) strategy.
This strategy initially learns semantics with accurate geometry using ground-truth depth, and then gradually mixes predicted depth to adapt the model to the predicted geometry.
Extensive experiments on the widely-used Occ3D-nuScenes benchmark demonstrate the superiority of our method, which achieves a 39.4 mIoU with 20.0 FPS.
This result is $\sim 3 \times$ faster and +1.9 mIoU higher compared to FB-OCC, the winner of CVPR2023 3D Occupancy Prediction Challenge.
Our code will be made open-source.

\end{abstract}

\begin{figure}[t]
    \centering
    \includegraphics[width=1.0\linewidth]{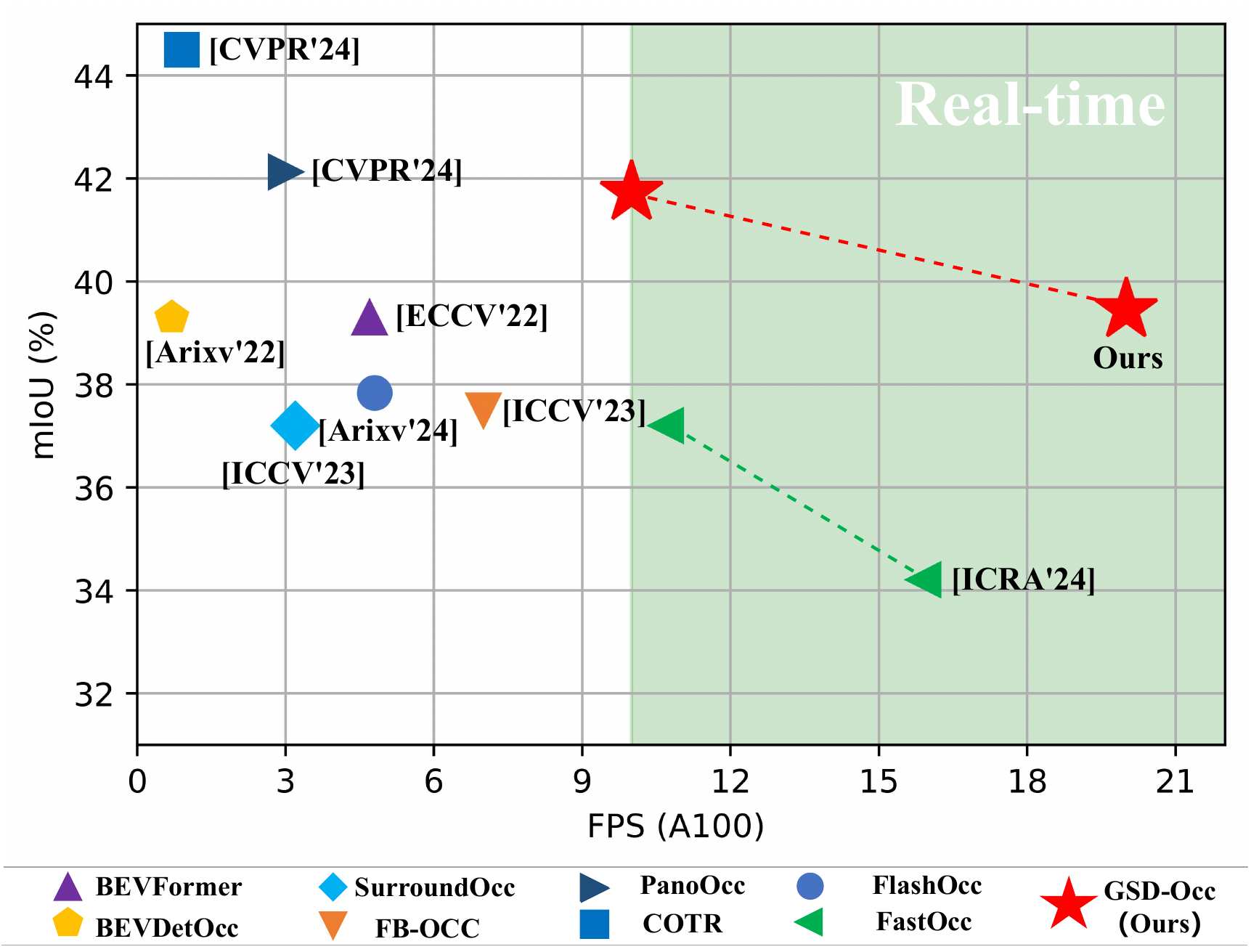}
    \caption{The inference speed (FPS) and accuracy (mIoU) of occupancy prediction methods on the Occ3D-nuScenes~\cite{occ3d} benchmark.
    GSD-Occ has a clear advantage of accuracy in real-time.}
    \label{fig:first}
\end{figure}

\section{Introduction}
\label{sec:intro}
Vision-based occupancy prediction~\cite{tesla} leverages surround-view camera images of ego vehicle to estimate object occupancy and semantics within a voxel space~\cite{tpvformer,occ3d,fb-occ,Surroundocc,sparseocc,COTR,Panoocc}.
Compared to 3D object detection~\cite{Bevfusion,center,Pointpillars}, it offers finer-grained 3D scene perception and produces a LiDAR-free alternative. 
Besides, by determining object presence within grid cells, occupancy prediction models can identify general objects, effectively handling out-of-vocabulary and unusual obstacles.

Despite these strengths, existing methods~\cite{bevstereo,Bevformer,Monoscene,Voxformer,OccFormer} often suffer from low inference speed (\textit{e.g.}, 1 $\sim$ 3 FPS on Nvidia A100~\cite{Bevdet4d,Panoocc,COTR}) and high memory usage (\textit{e.g.}, $>$ 10,000 MB~\cite{Panoocc,COTR}) due to the high computational cost of 3D voxel features.
These limitations hinder their application in AD vehicles equipped with on-board GPUs.
To redirect the focus from solely prioritizing accuracy to also considering deployment friendliness, we conduct an extensive evaluation of the speed and memory usage for most public available methods.

\begin{figure}[t]
    \centering
    \includegraphics[width=0.9\linewidth]{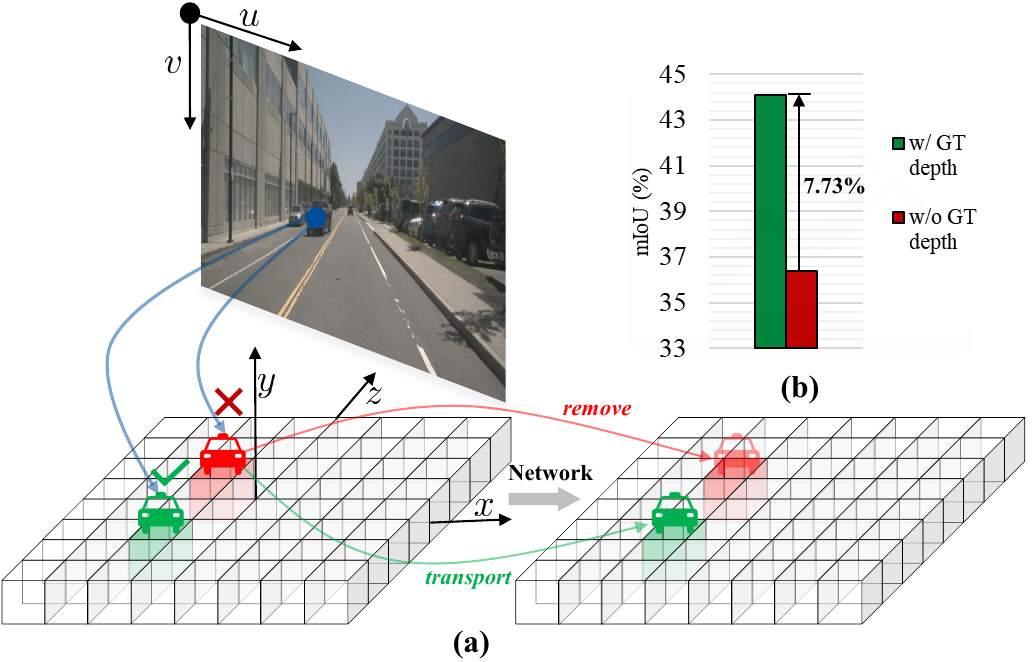}
    \caption{Illustration of the geometric-semantic coupling problem. 
    (a) Incorrect prediction depth can result in inaccurate 2D-to-3D feature projection, which requires refinement and correction by the subsequent network. (b) illustrates the performance gap between using prediction depth and ground-truth depth, which further underscores the importance of addressing this issue.
    }
    \label{fig:motivation}
\end{figure}

Through an extensive review and evaluation of existing methods, we identify a core challenge in achieving both fast and accurate performance: \textbf{\textit{the strong coupling between geometry and semantic}}.
As shown in Fig.~\ref{fig:motivation}, the geometric prediction (depth) serves as the input of the 2D-to-3D feature projection and impacts the downstream semantic classification.
Therefore, inaccurate prediction depth can destroy the discriminative power of features and increases optimization difficulty.
To address this issue, we propose to decouple geometric and semantic learning from both network design and learning strategy two perspectives.


As for the network design, existing methods primarily rely on \textit{heavy} 3D networks~\cite{Bevdet4d,COTR} to simultaneously refine geometric structure and learn semantic knowledge.
However, the high computational cost of 3D networks is unaffordable for real-time methods.
Recently, several works~\cite{FlashOcc,FastOcc} collapse 3D voxel features into BEV features to 
improve efficiency, but they often fail to achieve satisfactory accuracy (\textit{e.g.}, FastOcc~\cite{FastOcc} in Fig.~\ref{fig:first}), as the BEV representation loses height information~\cite{COTR}.
Therefore, it is both natural and promising to adopt a hybrid BEV-Voxel representation, which combines the strengths of computational efficiency in BEV representation and geometric integrity in voxel representation.
To this end, we propose a Geometric-Semantic Dual-Branch Network (GSDBN) guided by two principles: \textbf{\textit{sparse geometry}} and \textbf{\textit{dense semantics}}. 
In the BEV branch, we employ BEV-level temporal fusion and a 2D semantic encoder with U-Net~\cite{U-net} structure to extract dense semantic features.
In the voxel branch, we propose a 3D geometric encoder with a re-parameterized 3D large-kernel convolution, which refines the sparse geometric structure with enhanced receptive field and reduces computation through the re-parameterization technique.
To fuse the features of two branches, we propose a BEV-Voxel lifting module, which projects BEV-level semantic features into the voxel space along the height dimension, thus effectively recovering the lost height information.

As for the learning strategy, followed by Lift-Splat-Shoot (LSS)~\cite{lss}, almost all existing methods~\cite{Bevdet4d,FlashOcc,FastOcc,COTR} directly utilize the prediction depth for 2D-to-3D view transformation. However, they overlook that the prediction depth is not always accurate, especially at the early stage of training, which exacerbates the coupling problem and leads to unstable optimization.
Inspired by language models~\cite{radford2018improving,radford2019language,brown2020language}, which provide sequential ground-truth tokens to predict the next token, we replace the prediction depth with the ground-truth depth for 2D-to-3D view transformation during training.
However, this strategy performs poorly when using the prediction depth for testing, as the model is not adapted to the prediction depth and cannot correct errors in the predicted geometry.
To this end, we introduce a Geometric-Semantic Decoupled Learning (GSDL) strategy.
Initially, we use ground-truth depth for 2D-to-3D view transformation to maintain accurate geometric structure, allowing for isolated semantic learning.
Gradually, we mix the ground-truth depth with the prediction depth, which enables the model to learn to refine the predicted geometry. 
By decoupling the learning of geometric refinement and semantic knowledge, we effectively reduce the optimization difficulty and achieve further accuracy improvements without incurring additional deployment costs.

Our contributions can be summarized as follows:

\begin{itemize}
    \item We conduct an extensive evaluation of speed and memory usage for most public available methods, which aims to redirect the focus from solely prioritizing accuracy to also considering deployment friendliness.
    \item We propose a dual-branch network with a hybrid BEV-Voxel representation, which separates the learning of sparse geometry and dense semantics, ensuring both computational efficiency and geometric integrity.
    \item We propose a novel learning strategy to decouple the learning of geometric refinement and semantic knowledge, which achieves consistent accuracy improvements across various pre-training models and methods.
    \item We propose GSD-Occ, a Geometric-Semantic Disentangled Occupancy predictor, which establishes a new state-of-the-art with 39.4 mIoU and 20.0 FPS for real-time occupancy prediction.
\end{itemize}

\section{Related works}
\label{sec:related_works}

\condenseparagraph{Vision-based BEV Perception.} 
Bird's-Eye View (BEV) perception~\cite{li2022delving} has recently seen significant advancements, developing as a crucial component in autonomous driving (AD) due to its computational efficiency and rich visual semantics.
By leveraging 2D-to-3D view transformation to project camera image features into the BEV representation, multiple tasks can be integrated into a unified framework.
View transformation can be broadly categorized into two types: forward projection and backward projection. 
The former employs explicit depth estimation to project image features into 3D space~\cite{lss,bevdet,Bevdepth,bevstereo,Bevdet4d}.
In contrast, the latter first initializes a BEV space and then implicitly models depth information by querying image features using a spatial cross-attention~\cite{ddetr,wang2022detr3d,Bevformer,bevformerv2,jiang2023polarformer}.
Although BEV perception excels in 3D object detection, it still struggle with corner-case and out-of-vocabulary objects, which are crucial for ensuring the safety of autonomous driving.  
To address this issue, 3D occupancy prediction has been proposed, quickly emerging as a promising solution in AD~\cite{tesla}.

\condenseparagraph{3D Occupancy Prediction.} 
3D occupancy prediction reconstructs the 3D space using continuous voxel grids, which offers an enhanced geometry information and capability in detecting general objects.
A straightforward idea is to replace the BEV representation of 3D object detection methods with the voxel representation, and then append a segmentation head~\cite{bevdet,Bevformer,occ3d}.
However, voxel representations incur substantial computational and memory costs compared to BEV.
To address this, TPVFormer~\cite{tpvformer} divided the 3D space into three-view planes for feature extraction, followed by interpolation to recover voxel representations.
SurroundOcc~\cite{Surroundocc} and CTF-Occ~\cite{occ3d} utilized multi-scale encoders to gradually enhance voxel representations.
FB-OCC~\cite{fb-occ} adapt a hybrid of forward and backward view transformation to complete sparse voxel features.
COTR~\cite{COTR} proposes a compact voxel representation through downsampling, yet its feature enhancement network is so heavy that slows down the runtime significantly.
PannoOcc~\cite{Panoocc} introduced a novel panoramic segmentation task based on occupancy representation and adapt sparse 3D convolutions to decrease computation. 
Despite progress in accuracy, existing methods often suffer from speed and memory limitations. 
Therefore, we establish a benchmark that incorporates speed and memory usage to provide a more comprehensive  and fair assessment of methods.

\condenseparagraph{Deployment-Friendly Occupancy Prediction.} 
Recently, several works have focused on the deployment friendliness of occupancy prediction.
For example, FlashOcc~\cite{FlashOcc} directly uses a BEV representation to predict geometry and semantic, thereby reducing computational costs.
Similarly, FastOcc~\cite{FastOcc} employed a BEV representation but enhanced it using a residual structure that integrates voxel features obtained from view transformation.   
SparseOcc~\cite{sparseocc} employed a pure sparse transformer-based network to reduce computation.
However, these methods typically evaluate the speed or memory usage of only a limited set of methods.
To establish a comprehensive and fair evaluation benchmark, this work evaluates most public available methods using a same experimental environment.
Moreover, while existing methods significantly improve efficiency, they often fail to achieve satisfactory accuracy in real-time.
This work addresses this limitation by decoupling the learning of geometry and semantic, thereby achieving both real-time and accurate performance.

\begin{figure*}[t]
    \centerline{\includegraphics[width=0.95\linewidth]{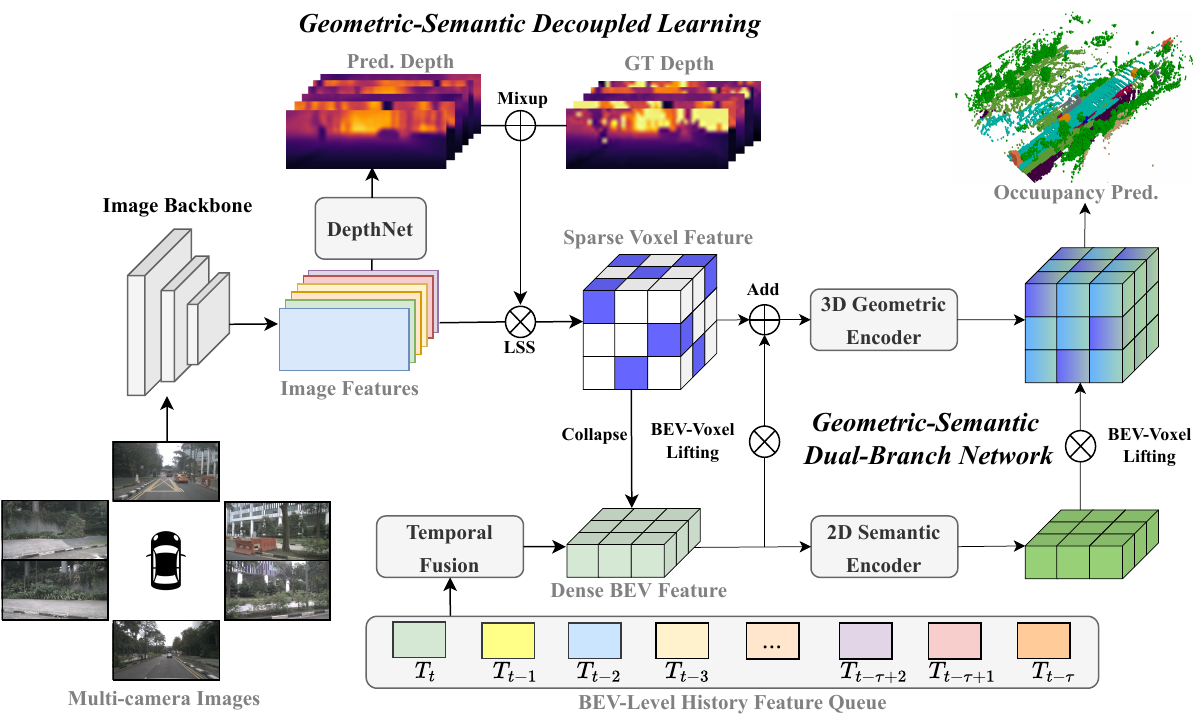}}
    \caption{
        The overview of GSD-Occ. Multi-camera images are first fed into an image backbone network to get image features, and DepthNet~\cite{Bevdepth} is used to predict a depth distribution.
        The Lift-Splat-Shoot (LSS)~\cite{lss} module is then employed to explicitly transform 2D image features into 3D voxel features.
        Subsequently, the geometric-semantic dual-branch network exploits a hybrid BEV-Voxel representation to efficiently maintain geometric structure while extracting rich semantics.
        The geometric-semantic decoupled learning strategy injects ground-truth depth into LSS to separate the learning of geometric correction and semantic knowledge, thereby further improving accuracy.
    } \label{fig:overview}
    \vspace{-0.5em}
\end{figure*}

\begin{figure*}[t]
    \vspace{-0.3em}
    \centerline{\includegraphics[width=0.9\linewidth]{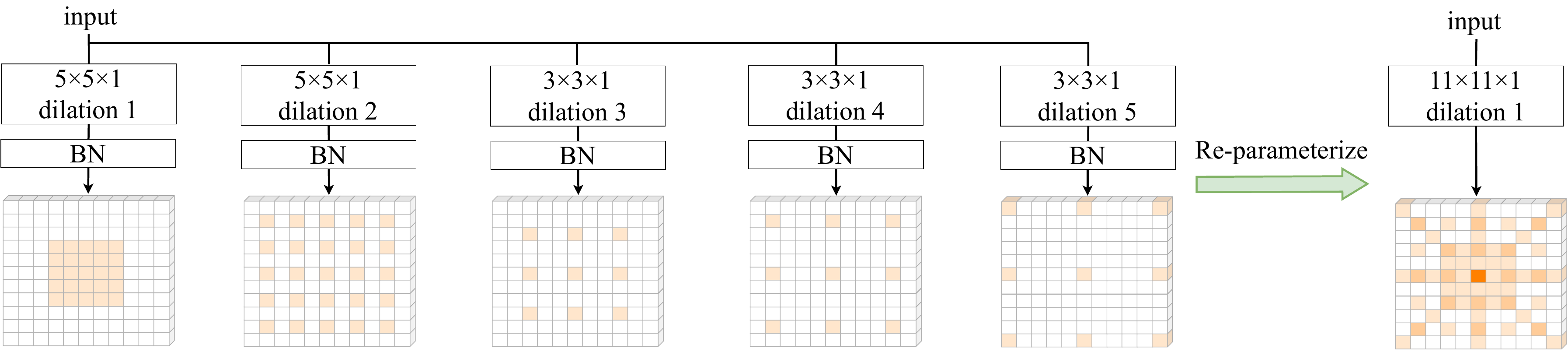}}
    \caption{
    Illustration of the large-kernel 3D convolutional re-parameterization technique in 3D geometric encoder. It uses parallel dilated small-kernel 3D convolutions to enhance a non-dilated large-kernel 3D convolution.
     This example shows $[K_H,K_W,K_Z]=[11,11,1]$.
    } \label{fig:3dencoder}
    \vspace{-0.5em}
\end{figure*}

\section{Method}
\label{sec:method}

\subsection{Problem Formulation}

Given a sequence of images $I_{i,t}\in\mathbb{R}^{H_i \times W_i \times 3}$ from $N_c$ surround-view cameras over $T$ frames, where $i\in\{1,..., N_c\}$ and $t\in\{1,...,T\}$.
The camera intrinsic parameters$\{K_i\}$ and extrinsic parameters$\left\{\left[R_i \mid t_i\right]\right\}$ in each frame are also known.
Vision-based 3D occupancy prediction aims to estimate the state of 3D voxels within the range $[X_s, Y_s, Z_s, X_e, Y_e, Z_e]$ around the ego vehicle.
The shape of the 3D voxels is $[X, Y, Z]$ (\textit{e.g.}, [200,200,16] in~\cite{occ3d}), where $[\frac{X_e-X_s}{X}, \frac{Y_e-Y_s}{Y}, \frac{Z_e-Z_s}{Z}]$ is the size of each voxel.
Each voxel contains \textit{occupancy state} (``occupied'' or ``empty'') and specific \textit{semantic information} (``category'' or ``unknown'').
Benifit from the learning of occupancy,  3D occupancy prediction can develop a general object representation to handle out-of-vocabulary and unusual obstacles.

\subsection{Overall Architecture}
The overview of the Geometric-Semantic Disentangled Occupancy predictor (GSD-Occ) is shown in Fig.~\ref{fig:overview}, which includes an image encoder to extract image features, a 2D-to-3D view transformation to project image features into 3D space, a geometric-semantic dual-branch network (Sec.~\ref{sec:GSDBN}) to efficiently maintain geometric integrity and extract rich semantics, and a geometric-semantic decoupled learning strategy (Sec.~\ref{sec:GSDL}) to further enhance the ability of geometric refinement and semantic learning.

\textbf{Image Encoder.} Given a set of surround-view camera images at $T$-th frame, denoted as $I_T = \{I_{i,T}\in\mathbb{R}^{H_i \times W_i \times 3}\}_{i=1}^{N_c}$, we employ a pre-trained backbone network (\textit{e.g.}, ResNet-50~\cite{he2016deep}) to extract image features $F=\{F_i \in \mathbb{R}^{C_F \times H_F \times W_F}\}_{i=1}^{N_c}$, where $[H_i, W_i, 3]$ and $[C_F,H_F,W_F]$ are the shapes of images and features, respectively.
$N_c$ is the number of cameras on the ego-vehicle.  

\textbf{2D-to-3D View Transformation.}
2D-to-3D view transformation aims to convert 2D image features $F$ to voxel representation.
Given the limited learning capacity of real-time models, we adopt an explicit view transformation module~\cite{lss} supervised by depth.
Specifically, the image features $F$ are first fed into the DepthNet~\cite{Bevdepth} to generate a predicted depth distribution $D=\left\{D_i \in \mathbb{R}^{D_{b i n} \times H_F \times W_F}\right\}_{i=1}^{N_c}$, where $D_{b i n}$ is the number of depth bins.
With $F$ and the $D$ as input, a pseudo point cloud feature $P \in \mathbb{R}^{N_c D_{b i n} H_F W_F \times C_p}$ can be obtained through outer product $F \otimes D$.
Finally, voxel-pooling is applied to the $P$ to obtain the voxel features $V \in\mathbb{R}^{C \times \frac{X}{2} \times \frac{Y}{2} \times \frac{Z}{2}}$, with 2$\times$ downsampling to reduce computational complexity.

\subsection{Geometric-Semantic Dual-Branch Network}
\label{sec:GSDBN}
The key idea behind Geometric-Semantic Dual-Branch Network (GSDBN) module is to employ a hybrid BEV-Voxel representation, where sparse voxel features server as \textit{``skeleton''} to maintain 3D geometric information and computation-efficient BEV features are used as  \textit{``flesh''} to complete voxel features with semantic information.
We first elaborate the two principles for the design of GSDBN \textit{i.e.}, \textbf{\textit{sparse geometry}} and \textbf{\textit{dense semantic}}.
(1) Sparse geometry in 3D occupancy grids reflects the discretization of the physical world, which leads to the sparsity of voxel features, with over 35\% of values being zero after the 2D-to-3D view transformation. 
(2) Dense semantic, on the other hand, is necessary to maintain the model's classification ability, as excessive zero values can severely degrade performance.
Then, we detail GSDBN based on the two key principles.

\subsubsection{Semantic BEV Branch}
\textbf{BEV-Level Temporal Fusion.}
To reduce computation and memory costs, we propose using BEV features instead of voxel features employed in ~\cite{fb-occ} for temporal fusion.
Besides, we introduce a history feature queue as in~\cite{park2022time} to avoid time-consuming and redundant feature re-computation in ~\cite{FlashOcc,Bevdet4d,COTR}.
Specifically, we collapse the voxel feature $V$ along the height dimension to obtain the BEV feature $B\in\mathbb{R}^{C\times \frac{X}{2} \times \frac{Y}{2}}$, and maintain a memory queue of length $\tau$ to store the historical BEV features.
To fuse the BEV features of the historical $\tau$ frames with the current frame, we first warp them to the current timestamp $T$ and then feed them into 2D convolutions to obtain the temporal BEV features $B_t\in\mathbb{R}^{C \times \frac{X}{2} \times \frac{Y}{2}}$.  
The sparsity of voxel features enable BEV features  to retain rich information, resulting in an acceptable accuracy degradation (0.69 mIoU) and a notable decrease in inference time (0.025 s).

\textbf{2D Semantic Encoder.}
We employ a light-weight 2D UNet-like~\cite{ronneberger2015u} encoder to extract features with rich semantic information.
Specifically, the temporal BEV feature $B_t$ is downsampled and then upsampled by a factor of 4, with residuals utilized to fuse multi-scale features. 
This process yields the semantic BEV features $B_s\in\mathbb{R}^{C^{\prime} \times \frac{X}{2} \times \frac{Y}{2}}$.

\subsubsection{Geometric Voxel Branch}
\textbf{3D Geometric Encoder.}
Inspired by ~\cite{Repvgg,unireplknet}, we extend re-parameterization technique to 3D occupancy prediction by designing a large-kernel re-parameterized 3D convolution for geometric encoding.
By this way, we can enhance the receptive field of voxel features to refine the geometric structure, while the re-parameterization technique significantly reduces inference time.

During training, we employ a non-dilated small-kernel and multiple dilated small-kernel 3D convolutions along with batchnorm (BN) layers.
This combination helps capture small-scale patterns and enhance the receptive filed.
During inference, these parallel small-kernel 3D convolutions can be converted into a large-kernel convolution to improve efficiency.

As illustrated in Fig~\ref{fig:3dencoder}, we show a case of a 3D convolutional kernel with size $[K_X,K_Y,K_Z]$ equals to $[11,11,1]$.
Since omitting pixels in the input is equivalent to inserting extra zero entries into the convolution, a dilated convolution with a small kernel can be equivalently converted into a non-dilated one with a sparse larger kernel~\cite{unireplknet}.
For a small 3D convolutional kernel $W \in \mathbb{R}^{k_x \times k_y \times k_z}$ with the dilation rate $(r_x,r_y,r_z)$, this transformation can be elegantly implemented by a transpose convolution:
\begin{equation}
    W^{\prime} = \mathrm{conv\_transpose3d}(W, I, s=(r_x,r_y,r_z))
\end{equation}where $I\in \mathbb{R}^{1 \times 1 \times 1}$ and $s$ means the stride.
Then, the sparse kernel $W^{\prime}$ and the subsequent 3D BN layer (with the parameters of accumulated mean $\mu$, standard deviation $\sigma$, the learned scaling factor $\gamma$, and the learned bias $\beta$) can be converted into a convolution with a bias vector:
\begin{equation}
    W^{\prime \prime} = \frac{\gamma}{\sigma} W^{\prime}, \quad b^{\prime \prime} = - \frac{\mu \gamma}{\sigma} + \beta.
\end{equation}
The weight and bias of the final large kernel can be obtained by summing $W^{\prime \prime}$ and  $b^{\prime \prime}$ across multiple parallel convolutions:
\begin{equation}
    \hat{W} = \sum_{i=1}^{C_s} \mathrm{zero\_padding} (W^{\prime \prime}_i), \quad \hat{b} = \sum_{i=1}^{C_s}(b^{\prime \prime}_i),
\end{equation}where $C_s$ is the number of small-kernel convolutions and $\mathrm{zero\_padding}$ is the zero-padding function that pads $W^{\prime \prime}$ to the size of large kernel $[K_X,K_Y,K_Z]$.
Finally, the geometric voxel features $V_g \in \mathbb{R}^{C^{\prime} \times \frac{X}{2} \times \frac{Y}{2} \times \frac{Z}{2}}$ are obtained by performing the 3D convolution with the weight
$\hat{W}$ and bias $\hat{b}$ of the large kernel.

\textbf{BEV-Voxel Lifting Module.}
To fuse the output of BEV and voxel branches, we propose a BEV-Voxel lifting (BVL) module that projects BEV features into voxel space.
This design is inspired by LSS~\cite{lss}, but it projects BEV features along the height dimension instead of image features along the depth dimension.
As shown in Fig.~\ref{fig:overview}, the BVL module is applied to the temporal BEV feature $B_t$ and the semantic BEV feature $B_s$.
For example, using $B_s$ as input, a context branch generates height-aware features $B_s^{\prime} \in \mathbb{R}^{C^{\prime} \times \frac{X}{2} \times \frac{Y}{2}}$, while a height branch predicts a height distribution $H^{\prime} \in \mathbb{R}^{\frac{X}{2} \times \frac{Y}{2} \times \frac{Z}{2}}$.
Then, the semantic voxel features $V_s \in \mathbb{R}^{C^{\prime} \times \frac{X}{2} \times \frac{Y}{2} \times \frac{Z}{2}}$ are then obtained through the outer product $B_s\otimes H^{\prime}$.
Finally, the geometric-semantic decoupled features $V_{g\&s} \in \mathbb{R}^{C^{\prime} \times X \times Y \times Z}$ are obtained by summing the geometric voxel feature $V_g$ and the semantic voxel feature$V_s$, followed by upsampling 2 $\times$ using transpose 3D convolutions: $V_{g\&s} = \mathrm{upsample} (V_g + V_s)$.

\subsection{Geometric-Semantic Decoupled Learning}
\label{sec:GSDL}
In Sec.~\ref{sec:GSDBN}, the GSDBN module effectively mitigates the coupling problem between geometry and semantic through a dual-branch network design.
In this section, we further think about this issue from a learning perspective.
We focus on a key component for 2D-to-3D view transformation, \textit{i.e.}, the LSS module, which projects image features into voxel space by predicting a depth distribution. 
However, as the prediction depth is not always accurate, especially at the early stage of training, which would exacerbate the coupling problem and lead to unstable optimization.

An intuitive idea is to directly replace the prediction depth with the ground-truth depth during training in LSS, while using the prediction depth in inference.
This strategy is inspired by language models~\cite{radford2018improving,radford2019language,brown2020language}, where sequential ground-truth tokens are provided to predict the next token during training, but complete sentences are predicted in inference. 
However, this strategy performs poorly because the model does not learn how to refine the predicted geometry.


To address this issue, we propose a geometric-semantic decoupled learning (GSDL) strategy.
Specifically, we introduce ground-truth depth $\hat{D}=\{\hat{D}_i \in \mathbb{R}^{D_{b i n} \times H_F \times W_F}\}_{i=1}^{N_c}$ to LSS at the beginning of training, so that the model can separately focus on learning semantics with accurate ground-truth geometry.
Subsequently, we gradually mix the ground-truth depth $\hat{D}$ with the prediction depth $D$ during training to  
adapt the model to the predicted geometry.
The mixup depth $D^m$ can be obtained by conducting the arithmetic mean, using a factor $\alpha \in [0,1]$:
\begin{equation}
\begin{aligned}
    D^m &= \{D^m_i\}_{i=1}^{N_c}, \\
    D^m_i &= D_i \alpha + \hat{D}_i (1-\alpha).
\end{aligned}
\end{equation}
The value of $\alpha$ is determined by a projection function, which is monotonically increasing with respect to the number of training iterations.
We first transform the range of iterations from $x\in[0, T_{max}]$ to $x\in[-N_{\alpha}, N_{\alpha}]$, where $T_{max}$ is the maximum number of training iterations and $N_{\alpha}$ is a constant set to 5 in this work without careful selection.
We then employ a sigmoid function to smooth the training process:
\begin{equation}
        \alpha = \frac{1}{1+e^{r x}}
\label{eq:mixup}
\end{equation}where $r$ is a parameter that controls the steepness of the mixup.
As $\alpha \to 1$ by the end of training, the model gains the ability to refine predicted geometry and no longer requires ground-truth depth in inference.

\begin{table*}[t]
	\footnotesize
    \setlength{\abovecaptionskip}{0pt}
	\setlength{\tabcolsep}{0.02\linewidth}
	\newcommand{\classfreq}[1]{{~\tiny(\nuscenesfreq{#1}\%)}}  %
    \begin{center}
	\begin{tabular}{l|c|c|c|c|ccc}
		\toprule
		Method
		& Venue & Backbone & Image Size & Visible Mask & mIoU (\%) & FPS & Memory (MB)
		\\
		\midrule
        MonoScene~\cite{Monoscene} & CVPR'{\color{blue}22} & ResNet-101 & $928\times600$ & \ding{56} & 6.1 & - & -  \\
        OccFormer~\cite{OccFormer} & ICCV'{\color{blue}23} & ResNet-50 & $256\times704$ & \ding{56} & 20.4 &4.8 &7617 \\
        CTF-Occ~\cite{occ3d} & arXiv'{\color{blue}23} & ResNet-101 & $928\times600$  & \ding{56} &  28.5 & - & -  \\
        SparseOcc~\cite{sparseocc} & CVPR'{\color{blue}24} & ResNet-50 & $256\times704$ & \ding{56} & 30.6 & 17.7 & 6883 \\
        \rowcolor[HTML]{EFEFEF}
        GSD-Occ (Ours) & - & ResNet-50 & $256\times704$ & \ding{56} & \textbf{31.8} & \textbf{20.0} & \textbf{4759} \\
        \midrule
        BEVFormer~\cite{Bevformer} & ECCV'{\color{blue}22} & ResNet-101& $900\times1600$ &  \ding{52}& 39.3 & 4.7 & 6651 \\
        BEVDet4D~\cite{Bevdet4d} & arXiv'{\color{blue}22} & ResNet-50 & $256\times704$ & \ding{52} & 39.2 & 0.8 &6053 \\
        SurroundOcc~\cite{Surroundocc}&ICCV'{\color{blue}23}& ResNet-101 & $900\times1600$ & \ding{52} &37.1&3.2&5491\\
        FB-OCC~\cite{fb-occ}&ICCV'{\color{blue}23}&ResNet-50 & $256\times704$ &\ding{52} &37.5&7.0&5467\\
        FlashOCC~\cite{FlashOcc}&arXiv'{\color{blue}24}& ResNet-50&$256\times704$ &\ding{52} &37.8&4.8&\textbf{3143}\\
        FastOCC~\cite{FastOcc}&ICRA'{\color{blue}24}&ResNet-50&$320\times800$ &\ding{52} &34.2&15.9&-\\
        FastOCC*~\cite{FastOcc}&ICRA'{\color{blue}24}&ResNet-101&$320\times800$ &\ding{52} &37.2&10.7&-\\
        PanoOcc~\cite{Panoocc}&CVPR'{\color{blue}24}& ResNet-101&$900\times1600$ &  \ding{52}&42.1&3.0&11991\\
        COTR~\cite{COTR}&CVPR'{\color{blue}24}& ResNet-50&$256\times704$ &\ding{52} &\textbf{44.5}&0.9&10453\\
        \rowcolor[HTML]{EFEFEF}
        GSD-Occ (Ours) & - & ResNet-50 & $256\times704$ & \ding{52} & 39.4 & \textbf{20.0} & 4759 \\
        \rowcolor[HTML]{EFEFEF}
        GSD-Occ* (Ours) & - & ResNet-50 & $512\times1408$ & \ding{52} & 41.7 & 10.0 & 5185 \\
		\bottomrule
	\end{tabular}
    \end{center}
    \caption{\textbf{3D Occupancy prediction performance on the Occ3D-nuScenes dataset.} 
    The FPS of all methods are evaluated on an Nvidia A100 GPU, except for FastOCC, which is tested on an Nvidia V100 GPU as reported.
    ``-'' indicates that the metrics are not reported in the paper, and the code is also not open-source.
    Visible Mask refers to whether models are trained with visible masks.}
	\label{tab:nuscene_sota}
    \vspace{-1.5em}
\end{table*}

\begin{table*}[t]
	\footnotesize
    \setlength{\abovecaptionskip}{0pt}
	\setlength{\tabcolsep}{0.02\linewidth}
	\newcommand{\classfreq}[1]{{~\tiny(\nuscenesfreq{#1}\%)}}  %
    \begin{center}
	\begin{tabular}{l|c|c|c|c|ccc}
		\toprule
		Method & Venue & Backbone & Image Size & Epoch & RayIoU (\%) & FPS & Memory (MB)
		\\
		\midrule
        BEVFormer~\cite{Bevformer} & ECCV'{\color{blue}22} & ResNet-101 & $900\times1600$ & 24 & 32.4 & 4.7 & 6651  \\
        BEVDet4D~\cite{Bevdet4d} & arXiv'{\color{blue}22} & ResNet-50 & $256\times704$&90 & 29.6 & 0.8 &6053 \\
        FB-OCC~\cite{fb-occ}&ICCV'{\color{blue}23}&ResNet-50 & $256\times704$ &90 &33.5&7.0&5467\\
        SparseOcc~\cite{sparseocc} & ECCV'{\color{blue}24} & ResNet-50 & $256\times704$ & 24 & 34.0 & 17.7 & 6883 \\
        \rowcolor[HTML]{EFEFEF}
        GSD-Occ (Ours) & - & ResNet-50 & $256\times704$ & 24 & \textbf{38.9} & \textbf{20.0} & \textbf{4759} \\
		\bottomrule
	\end{tabular}
    \end{center}
    \caption{\textbf{3D Occupancy prediction performance on the Occ3D-nuScenes dataset using the RayIoU metric proposed by~\cite{sparseocc}.} 
    The FPS of all methods are evaluated on an Nvidia A100 GPU.}
	\label{tab:rayiou}
    \vspace{-1.5em}
\end{table*}

\section{Experiments}
\label{sec:experiments}

\begin{figure*}[t]
    \centerline{\includegraphics[width=0.95\linewidth]{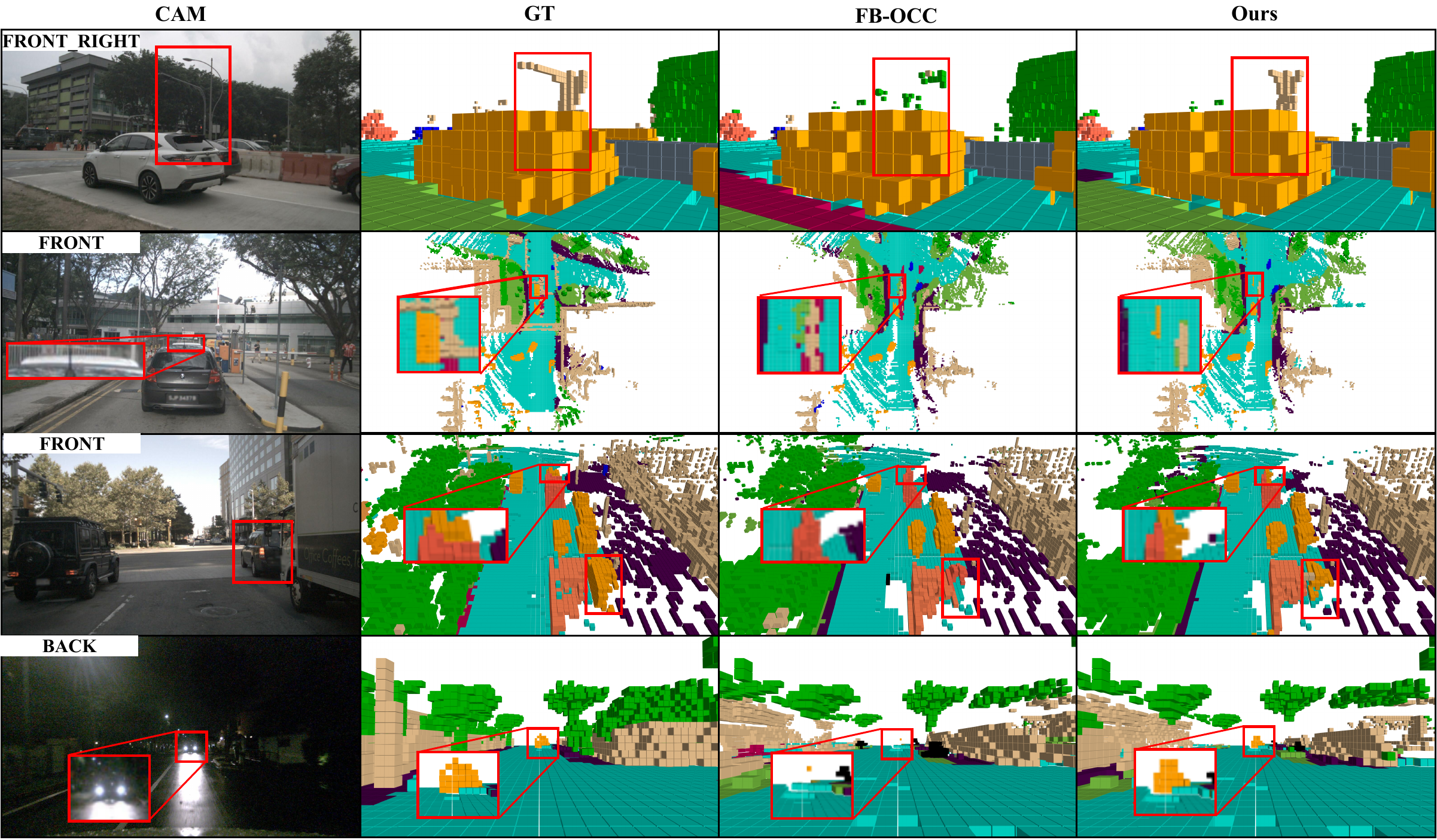}}
    \caption{\textbf{Qualitative results comparison between FB-OCC and our method.} 
    The results demonstrate that our method is able to construct more detailed geometry (Row~1 and Row~2), more accurate semantics (Row~3), and stronger adaptability in night (Row~4).
    }
    \label{fig:vis_results}
    \vspace{-2mm}
\end{figure*}


\begin{table}[t]
    \begin{center}
    \scalebox{0.8}{
    \begin{tabular}{cccc|cc}
    \toprule
    \multicolumn{4}{c|}{GSDBN} & \multirow{2}{*}{mIoU} & \multirow{2}{*}{FPS}\\
    \cline{1-4}
    \makecell{2D \\ Encoder} & \makecell{Temporal \\ Fusion} & \makecell{3D \\ Encoder} & BVL &  & \\
    \midrule
     & & & & 35.11 & 27.0\\
    \ding{52} & & & & 36.38 & 23.1 \\
    \ding{52} & 3D & & & 39.09 & 13.9 \\
    \ding{52} & 2D & & & 38.40 & 21.4 \\
    \rowcolor[HTML]{EFEFEF}
    \ding{52} & 2D & \ding{52} & \ding{52} & 38.90 & 20.0 \\
    \bottomrule
    \end{tabular}}
    \end{center}
    \setlength{\abovecaptionskip}{0pt}
    \caption{\textbf{Ablation study on each component of GSDBN.} ``3D'' and ``2D'' denote conducting temporal fusion with voxel or BEV features. ``BVL'' refers to the BEV-Voxel Lifting module.}
    \label{tab:ablation_compo}
    \vspace{-1.75em}
\end{table}

\begin{table}[t]
    \begin{center}
    \scalebox{0.85}{
    \begin{tabular}{cccc}
    \toprule
    Method & \makecell{Pretrained \\ model} & GSDL & mIoU \\
    \midrule
    FB-OCC~\cite{fb-occ} & BEVDepth &  & 37.5 \\
    \rowcolor[HTML]{EFEFEF}
    FB-OCC~\cite{fb-occ}  & BEVDepth & \ding{52} & 37.82 (+0.32) \\
    \midrule
    GSD-Occ & ImageNet & &36.48\\
    \rowcolor[HTML]{EFEFEF}
    GSD-Occ & ImageNet & \ding{52} &36.88 (+0.40)\\
    \midrule
    GSD-Occ & BEVDepth & &38.90\\
    \rowcolor[HTML]{EFEFEF}
    GSD-Occ & BEVDepth & \ding{52} &39.45 (+0.55)\\
    \bottomrule
    \end{tabular}}
    \end{center}
    \setlength{\abovecaptionskip}{0pt}
    \caption{\textbf{Effectivenss of GSDL with different pretrained models and methods.} BEVDepth means the model weight in~\cite{Bevdepth}.}
    \label{tab:effect_gsdl}
\end{table}

\begin{table*}
\begin{floatrow}
\capbtabbox{
    \scalebox{0.85}{
    \vspace{-2mm}
    \begin{tabular}{ccc}
        \toprule
        Lifting Method & mIoU & FPS \\
        \midrule
        Channel-to-Height~\cite{FlashOcc} & 38.62 & 18.7 \\
        Repeat + 3D Conv~\cite{FastOcc} & 38.42 & 18.3 \\
        \rowcolor[HTML]{EFEFEF}
        BVL & 39.45 & 20.0 \\
        \bottomrule
        \end{tabular}}
}
{\caption{\textbf{The effectivenss of BVL.}}
    \label{tab:bvl}
}
\hspace{-0.5cm} 
\capbtabbox{
    \scalebox{0.85}{
        \vspace{-2mm}
        \begin{tabular}{ccc}
            \toprule
            \makecell{Number of \\ history frames} & mIoU & FPS \\
            \midrule
            1 (short) & 37.28 & 21.0 \\
            7 (moderate) & 38.49 & 20.6 \\
            \rowcolor[HTML]{EFEFEF}
            15 (long) & 39.45 & 20.0 \\
            31 (very long) & 39.32 & 19.3 \\
            \bottomrule
            \end{tabular}}
}
{\caption{\textbf{The impact of different number of history frames in temporal fusion.}}
        \label{tab:history_fusion}
}
\hspace{0.5cm} 
\capbtabbox{
    \scalebox{0.85}{ 
    \vspace{-2mm}
    \begin{tabular}{ccc}
        \toprule
        \makecell{Kernel size \\ of 3D convolution} & mIoU & FPS\\
        \midrule
        $7 \times 7 \times 1$ & 38.67 & 19.4/20.4 \\
        $9 \times 9 \times 1$ & 38.70 & 19.0/20.2 \\
        \rowcolor[HTML]{EFEFEF}
        $11 \times 11 \times 1$ & 38.90 & 18.6/20.0 \\
        $13 \times 13 \times 1$ & 38.65 & 18.2/19.6 \\
        $15 \times 15 \times 1$ & 38.74 & 17.9/19.4 \\
        \bottomrule
        \end{tabular}}
}
{\caption{\textbf{The impact of different kernel sizes in 3D encoder.} ``-/-'' denotes the FPS of before and after re-parameterization.}
    \label{tab:3dencoder}
}
\end{floatrow}
\end{table*}

\subsection{Experimental Setup}
We evaluate our model using the Occ3D-nuScenes~\cite{occ3d} benchmark, which is based on nuScenes~\cite{caesar2020nuscenes} dataset and was constructed for the CVPR2023 3D occupancy prediction challenge.
The dataset consists of 1000 videos, split into 700 for training, 150 for validation, and 150 for testing.
Each key frame of video contains a 32-beam LiDAR point cloud, six RGB images from surround-view cameras, and dense voxel-wise semantic occupancy annotations.
The perception range in 3D voxel space is [-40m, -40m, -1m, 40m, 40m, 5.4m], with each voxel sized [0.4m,0.4m,0.4m].
The voxels contain 18 categories, including 16 known object classes, an unknown object class labeled as ``others'', and an ``empty'' class.
Following previous works~\cite{occ3d,fb-occ,FlashOcc,FastOcc}, we use the mean intersection over union (mIoU) across all classes to evaluate accuracy.

\subsection{Implementation Details}
Adhering to common practices~\cite{fb-occ,sparseocc,COTR}, we adopt ResNet-50~\cite{he2016deep} as the image backbone. 
We maintain a memory queue of length 15 to store historical features and fuse temporal information with 16 frames.
For the large-kernel re-parameterized 3D convolution in the geometric encoder, we set the size of convolution kernel to [11, 11, 1].
The steepness parameter $r$ is set to 5 in geometric-semantic decoupled learning.
During training, we use a batch size of 32 on 8 Nvida A100 GPUs.
Unless otherwise specified, all models are trained for 24 epochs using the AdamW optimizer~\cite{loshchilov2017fixing} with a learning rate $1\times 10 ^{-4}$ and a weight decay of 0.05.
During inference, we use a batch size of 1 on a single Nvidia A100 GPU.
The FPS and memory metrics are tested using the mmdetection3d codebase~\cite{mmdet3d2020}.

\subsection{Main Results}
In Tab.~\ref{tab:nuscene_sota} and Fig.~\ref{fig:first}, we compare GSD-Occ with previous state-of-the-art (SOTA) methods on the validation split of Occ3D-nuScenes.
GSD-Occ demonstrates real-time inference speed and low memory usage while achieving accuracy comparable to or better than non-real-time methods, such as BEVFormer~\cite{Bevformer}, BEVDet4D~\cite{Bevdet4d}, SurroundOcc~\cite{Surroundocc}, and FlashOCC~\cite{FlashOcc}.
When compared with FB-Occ~\cite{fb-occ}, the winner of CVPR 2023 occupancy challenge, GSD-Occ is $\sim 3 \times$ faster and shows a 1.9\% mIoU improvement.
Compared to other real-time occupancy prediction methods, GSD-Occ achieves a notable 5.2\% higher mIoU with even faster speed than FastOCC~\cite{FastOcc}.
These results highlight the effectiveness of geometric-semantic disentanglement in our method.
When we increase the input image size of GSD-Occ to $2\times$, the mIoU further improved by 2.3\% without bells and whistles.
The inference speed decreases by $2 \times$, which indicates a nearly linear relationship between input size and inference speed. 
This property enables GSD-Occ to efficiently handle high-resolution images.
Compared to more recent methods, GSD-Occ* achieves only 0.4\% lower mIoU than PannoOcc~\cite{Panoocc}, but it is $\sim 3 \times$ faster and  uses only  $\sim$ 50\% of the memory.
Although COTR~\cite{COTR} achieves 2.8\% higher mIoU than GSD-Occ*, it is significantly slower (\textgreater $ 10 \times$). 
Additionally, we also report the RayIoU metric proposed by ~\cite{sparseocc} in Tab.~\ref{tab:rayiou}.
GSD-Occ achieves 4.9 \% higher mIoU with faster speed and lower memory usage when compared with the recent SOTA method, SparseOcc~\cite{sparseocc}.

We further provide qualitative results in Fig.~\ref{fig:vis_results}. 
Despite significantly reducing computation, our method can also effectively perceive geometric details (even with few clues in Row~2) and accurate semantics (Row~3). 
Additionally, our method also performs well under night conditions (Row~4).

\begin{figure}[t]
    \centering
    \includegraphics[width=0.9\linewidth]{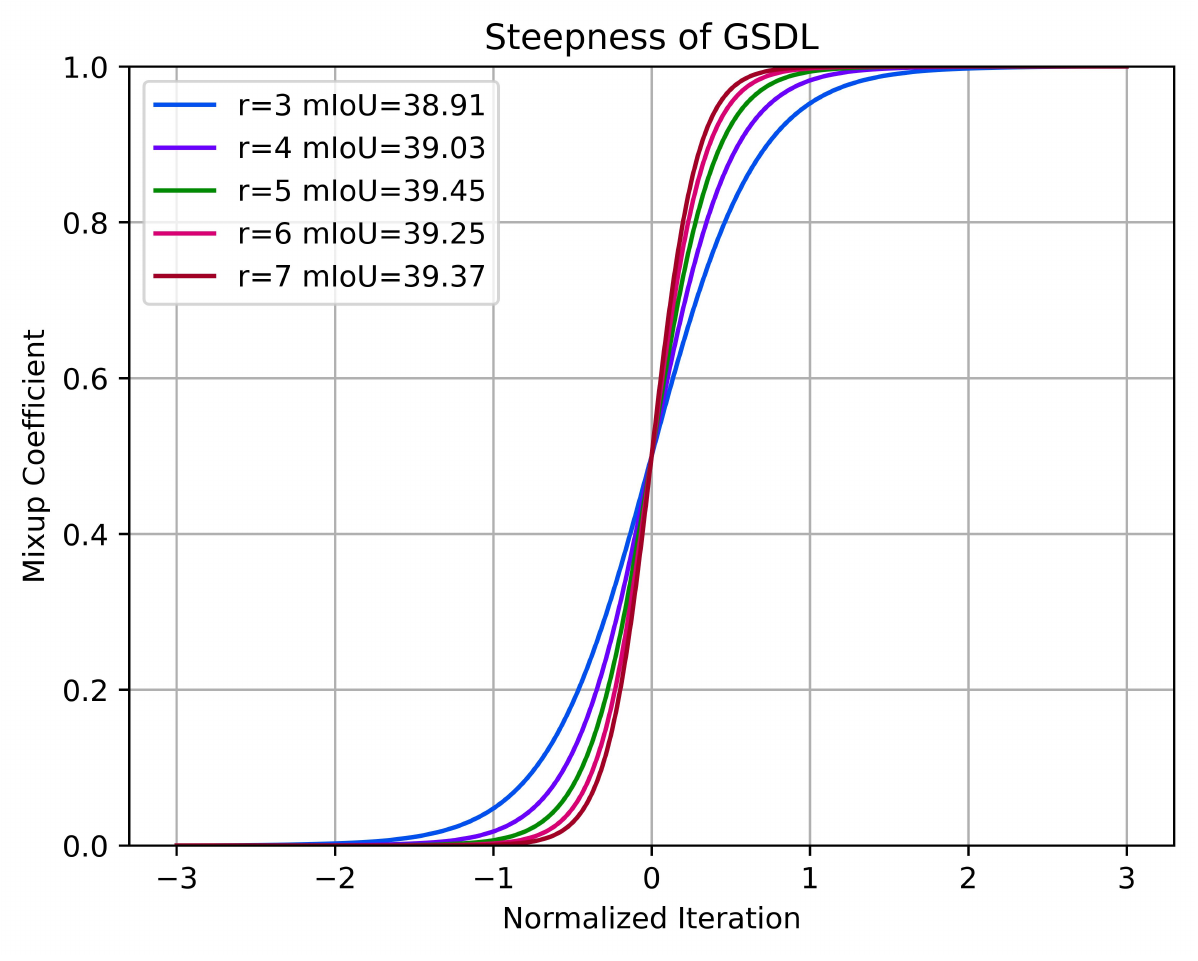}
    \caption{
        \textbf{Ablation study on the steepness (\textit{i.e.}, $r$ in Eq.~\ref{eq:mixup}) of GSDL.} ``Normalized iteration'' refers to the adjusted range of iterations as stated in ~\ref{sec:GSDL}.
        We show the steepness on different $r$ along with their corresponding performance.
    }
    \vspace{-3mm}
    \label{fig:gsdl}
\end{figure}

\subsection{Ablations}
In this section, we conduct conduct ablation experiments on validation split of Occ3d-nuScenes to delve into the effect of each module.

\subsubsection{Ablations on GSDBN}
The results are shown in Tab.~\ref{tab:ablation_compo}, we can observe that each component of geometric-semantic dual-branch network (GSDBN)  contributes to the overall performance.
The baseline model, which lacks temporal fusion and both 2D and 3D encoders, achieves fast speed (27.0 FPS) but falls short in accuracy (35.11\% mIoU).
For temporal fusion, although applying voxel features leads to 0.69 \% mIoU improvement when compared with using BEV features, it also introduces a significant inference delay (0.029s), which is costly relative to the accuracy gain.
Integrating the GSDBN module into the baseline model results in a 3.79\% mIoU improvement, with only a modest increase in computational cost (speed decreases from 27.0 FPS to 20.0 FPS). 
It demonstrates that GSDBN efficiently and effectively decouples the learning of geometry and semantic by a hybrid BEV-Voxel representation.

\subsubsection{Ablations on GSDL}

To prove the effectivenss of geometric-semantic decoupled learning (GSDL), we apply it to different pre-training models and methods, as shown in Tab.~\ref{tab:effect_gsdl}.
Without incurring additional computation costs, GSDL achieves consistent accuracy improvement across different pre-training models (BEVDepth~\cite{Bevdepth} and ImageNet~\cite{deng2009imagenet}) and methods (FB-OCC~\cite{fb-occ} and our GSD-Occ).
It highlights the generalizability of GSDL, which further decouples the geometry and semantic by a simple yet effective learning strategy.

\subsubsection{Additional Ablations}

\textbf{The Effectiveness of BVL.} 
We compare BEV-Voxel lifting (BVL) module with the other exisiting methods as shown in Tab.~\ref{tab:bvl}, it shows that BVL module achieves the best accuracy with the fastest speed, proving its effectiveness.

\textbf{Are More History Frames Better?} As illustrated in Tab.~\ref{tab:history_fusion}, we delve into the impact of various time-series lengths: short (1), moderate (7), long (15), and very long (31).
The results indicate that the long temporal fusion achieves the highest accuracy. 
Since we employ 2D temporal fusion with BEV features, the computational cost remains affordable even as the time-series length increases.

\textbf{Is a Larger 3D Convolutional Kernel Better?}
In Table~\ref{tab:3dencoder}, we present the results of different kernel sizes in 3D re-parameterized convolution. 
Adopting a kernel size of $11 \times 11 \times 1$ achieves the highest accuracy.
It indicates that correcting geometric errors requires a relatively large receptive field, but excessively large kernels can be counterproductive. 
Additionally, thanks to the re-parameterized technique we employed, the inference speed has significantly improved from 18.6 FPS to 20.0 FPS.

\textbf{Smooth or steep mixup of predicted and ground-truth depth?}
As shown in Fig.~\ref{fig:gsdl}, we plot the curve of Eq.\ref{eq:mixup} and conduct experiments to explore the impact of various steepness levels in GSDL.
When the steepness parameter $r$ is set to 5, we achieve the best accuracy.
This suggests that overly smooth mixup curves may hinder the model's ability to adapt to the predicted depth, while excessively steep curves can complicate the training process.

\section{Conclusion}
In this paper, we propose GSD-Occ, a employ-friendly real-time 3D occupancy prediction method that achieves accuracy comparable to many non-real-time methods.
To achieve this, we identify and address a core challenge: the strong coupling between geometry and semantic.
Specifically, we propose a geometric-semantic dual-branch network with a hybrid BEV-Voxel representation, which maintains both computational efficiency and geometric integrity.
Additionally, we propose a geometric-semantic decoupled learning strategy, which separates the learning of geometric correction and semantic knowledge, resulting in consistent accuracy improvements across various pre-training models and methods.
To validate the effectiveness of our method, we compare GSD-Occ with recent state-of-the-art (SOTA) methods on the Occ3D-nuScenes benchmark. 
The results demonstrate that GSD-Occ achieves new SOTA performance in real-time occupancy prediction.

{
    \small
    \bibliographystyle{ieeenat_fullname}
    \bibliography{main}
}

\end{document}